\documentclass[11pt]{article}

\usepackage{ACL2023}
\usepackage{times}
\usepackage{latexsym}

\usepackage{pdfpages}

\usepackage[T1]{fontenc}
\usepackage[utf8]{inputenc}

\usepackage{microtype}

\usepackage{inconsolata}

\usepackage{amsmath,amssymb,amsfonts}
\usepackage{algorithmic}
\usepackage{graphicx}
\usepackage{textcomp}
\usepackage{times}
\usepackage{latexsym}
\usepackage{graphicx}
\usepackage{subcaption}
\usepackage{svg}
\usepackage{amsmath}
\usepackage{lipsum}
\usepackage{enumitem}
\usepackage{microtype}

\title{Bangla Grammatical Error Detection Leveraging Transformer-based Token Classification}

\author{Shayekh Bin Islam, Ridwanul Hasan Tanvir, Sihat Afnan \\
        Department of Computer Science and Engineering,\\ BUET, Dhaka, Bangladesh \\
  \texttt{shayekh.bin.islam@gmail.com},  \texttt{1705016@ugrad.cse.buet.ac.bd},  \texttt{sihatafnan24@gmail.com}}

\begin{document}
% \nolinenumber
\maketitle
\begin{abstract}
% This document is a supplement to the general instructions for *ACL authors. It contains instructions for using the \LaTeX{} style file for ACL 2023.
% The document itself conforms to its own specifications, and is, therefore, an example of what your manuscript should look like.
% These instructions should be used both for papers submitted for review and for final versions of accepted papers.

% Recent advances in grammar chekcer mostly build for English language

Bangla is the seventh most spoken language by a total number of speakers in the world, and yet the development of an automated grammar checker in this language is an understudied problem. 
Bangla grammatical error detection 
is a task of detecting sub-strings of a Bangla text that contain grammatical, punctuation, or spelling errors, which  is crucial for developing an automated Bangla typing assistant. 

Our approach involves breaking down the task as a token classification problem and utilizing state-of-the-art transformer-based models.
Finally, we combine the output of these models and apply rule-based post-processing to generate a more reliable and comprehensive result. Our system is evaluated on a dataset consisting of over 25,000 texts from various sources. Our best model achieves a Levenshtein distance score of 1.04. Finally, we provide a detailed analysis
of different components of our system.

\end{abstract}

\section{Introduction}
% This document is a model and instructions for \LaTeX.
% Please observe the conference page limits. 
% Grammatical error detection (GED) is a well-known problem in NLP. 

Typing assistance has become increasingly important in today's digital age, especially with the rise of grammar-checking systems. Users expect typing assistance tools to not only detect and correct grammatical errors but also provide suggestions to improve their writing skills. With the help of predictive text, spell checkers, and auto-correct features, typing assistance tools can help users to type accurately and efficiently, reducing the time needed to correct errors. These features can also help users to expand their vocabulary and improve their grammar, leading to better writing skills and effective communication. With the rise of AI and machine learning, typing assistance tools such as Grammarly are becoming more advanced, helping users to improve their writing skills and communicate effectively.

Errors in Bangla text can be due to spelling, punctuation, or grammar. The spelling error itself can occur in various forms \cite{bijoy2022dpcspell} shown in Table~\ref{tab:spell}. Examples of errors similar to ERRANT \cite{bryant2017automatic} error classes are shown in Table~\ref{tab:errant}.  \cite{alam2020punctuation} approaches the punctuation correction problem only considering commas, periods, and question marks for ASR applications.

% \section{Background}

% Errors in Bangla text can be due to spelling, punctuation or grammar. The spelling error itself can occur in various forms \cite{bijoy2022dpcspell} shown in Table~\ref{tab:spell}. Examples of errors similar to ERRANT \cite{bryant2017automatic} error classes are shown in Table~\ref{tab:errant}. Alam et. al.\cite{alam2020punctuation} approaches the punctuation correction problem only considering commas, periods, and question marks for ASR applications.

\begin{table}[h]
  \centering
  \caption{Types of Spelling Errors}
  \label{tab:spell}
  \includegraphics[trim={50 200 100 50},clip,width=0.5\textwidth]{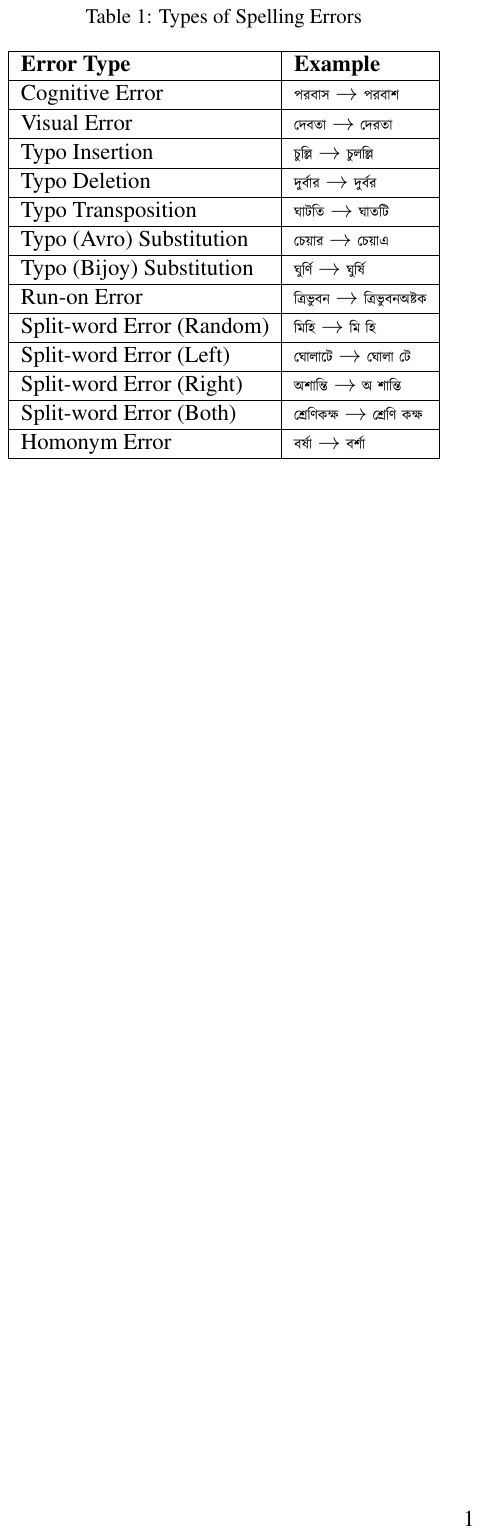}
  % \begin{tabular}{|l|l|}
    % \hline
    % \textbf{Error Type} & \textbf{Example} \\
    % \hline
    % Cognitive Error & {\bengali{পরবাস}}" $\rightarrow$\\\hline
    % Cognitive Error & {\beng পরবাস} $\rightarrow$ {\beng পরবাশ}\\\hline
    % Visual Error & {\beng দেবতা} $\rightarrow$ {\beng দেরতা}\\\hline
    % Typo Insertion & {\beng চুল্লি} $\rightarrow$ {\beng চুলল্লি}\\\hline
    % Typo Deletion & {\beng দুর্বার} $\rightarrow$ {\beng দুর্বর}\\\hline
    % Typo Transposition & {\beng ঘাটতি} $\rightarrow$ {\beng ঘাতটি}\\\hline
    % Typo (Avro) Substitution & {\beng চেয়ার} $\rightarrow$ {\beng চেয়াএ}\\\hline
    % Typo (Bijoy) Substitution & {\beng ঘুর্ণি} $\rightarrow$ {\beng ঘুর্ষি}\\\hline
    % Run-on Error & {\beng ত্রিভুবন} $\rightarrow$ {\beng ত্রিভুবনঅষ্টক}\\\hline
    % Split-word Error (Random) & {\beng মিহি} $\rightarrow$ {\beng মি হি}\\\hline
    % Split-word Error (Left) & {\beng ঘোলাটে} $\rightarrow$ {\beng ঘোলা টে}\\\hline
    % Split-word Error (Right) & {\beng অশান্তি} $\rightarrow$ {\beng অ শান্তি}\\\hline
    % Split-word Error (Both) & {\beng শ্রেণিকক্ষ} $\rightarrow$ {\beng শ্রেণি কক্ষ}\\\hline
    % Homonym Error & {\beng বর্ষা} $\rightarrow$ {\beng বর্শা}\\\hline    

  % \end{tabular}
  % \includepdf[pages=1, clip, trim=0pt 0pt 0pt 0pt]{table1.pdf}
\end{table}

\begin{table}[h]
  \centering
  \caption{Errors by ERRANT Classification}
  \label{tab:errant}
  \includegraphics[clip,width=0.5\textwidth]{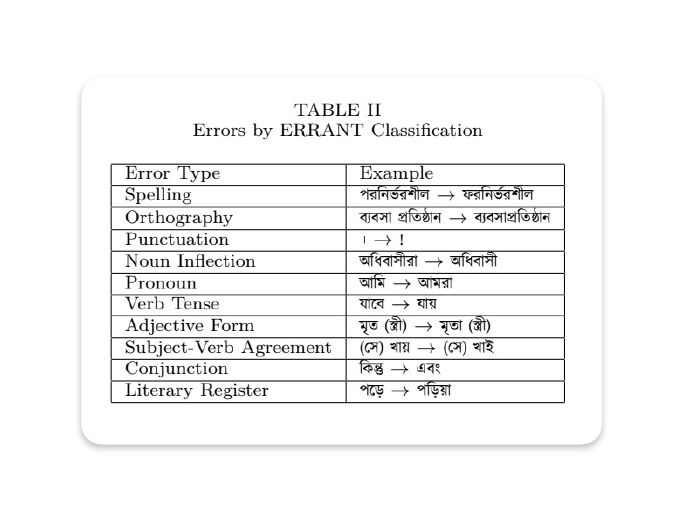}
  % \begin{tabular}{|l|l|}
    % \hline
    % \textbf{Error Type} & \textbf{Example} \\
    % \hline
    
    % Spelling & {\beng পরনির্ভরশীল} $\rightarrow$ {\beng ফরনির্ভরশীল}\\\hline
    % Orthography & {\beng ব্যবসা প্রতিষ্ঠান} $\rightarrow$ {\beng ব্যবসাপ্রতিষ্ঠান}\\\hline
    % Punctuation & {\beng ।} $\rightarrow$ {\beng !}\\\hline
    % Noun Inflection &  {\beng অধিবাসীরা} $\rightarrow$ {\beng অধিবাসী}\\\hline
    % Pronoun & {\beng আমি} $\rightarrow$ {\beng আমরা}\\\hline
    % Verb Tense & {\beng যাবে} $\rightarrow$ {\beng যায়}\\\hline
    % Adjective Form & {\beng মৃত (স্ত্রী)} $\rightarrow$ {\beng মৃতা (স্ত্রী)}\\\hline
    % Subject-Verb Agreement & {\beng (সে) খায়} $\rightarrow$ {\beng (সে) খাই}\\
    % \hline
    % Conjunction & {\beng কিন্তু} $\rightarrow$ {\beng এবং}\\\hline
    % Literary Register & {\beng পড়ে} $\rightarrow$ {\beng পড়িয়া}\\
    % \hline
    % Entry 4 & Entry 5 \\
    % \hline
  % \end{tabular}
\end{table}

\section{Related Works}

Numerous NLP techniques have been devised to address sentence-level errors, with statistical and rule-based methods being the most prevalent. Rule-based methods involve creating language-specific rules to tackle errors, while statistical approaches are favored for their language independence, making them more widely used than rule-based techniques.

In Bangla NLP, B. B. Chaudhuri has implemented an approximate string matching algorithm for detecting non-word errors \cite{chaudhuri2001reversed}, while N. UzZaman and M. Khan used a direct dictionary lookup method to handle misspelled word errors \cite{uzzaman2006comprehensive}, and Abdullah and Rahman employed it to detect typographical and cognitive phonetic errors \cite{abdullah2003generic}. P. Mandal and B. M. M. Hossain proposed a method based on the PAM clustering algorithm, which also did not address semantic errors \cite{mandal2017clustering}. A few works have been done at the semantic level. N. Hossain introduced a model that utilizes n-grams to check whether a word is correctly used in a sentence \cite{khan2014checking}, while K. M. Hasan, M. Hozaifa, and S. Dutta developed a rule-based method for detecting grammatical semantic errors in simple sentences \cite{hasan2014detection}.

The task of Bangla grammatical error detection is a span detection problem in general. One of the notable span detection problem is SemEval-2021 task 5: Toxic spans detection \cite{pavlopoulos2021semeval}, where the goal is to detect toxic spans within English passages. 

The solutions for the Toxic spans detection task formalize the task as a token-level sequence labeling (SL) problem \cite{zhu2021hitsz, nguyen2021s,wang2021hitmi,chen2021ynu,ghosh2021cisco, bansal2021iitk, chhablani2021nlrg}, as a span detection problem \cite{zhu2021hitsz, chhablani2021nlrg} and as a dependency parsing problem \cite{ghosh2021cisco}. Generating pseudo labels from external datasets \cite{nguyen2021s, bansal2021iitk} for semi-supervised learning is a notable part of some solutions. Here, the winning models are various transformer-based models along with LSTM and CRF. To combine the predictions of various models, they employ the union or the intersection of the predicted spans.

\section{Methodology}
% In this work, we follow the setup from \cite{chhablani2021nlrg} which was developed for solving the SemEval-2021 Task 5: Toxic Spans Detection \cite{pavlopoulos2021semeval}.

We formalize the task as a four-class token classification problem similar to the well-known BIO tagging scheme. Then, we employ transformer-based models along with LSTM and CRF. Next, we ensemble and post-process the model predictions to generate the final output.

\subsection{Data Pre-processing}
\subsubsection{Normalization}
Following the normalization scheme of BanglaBERT, we normalize punctuations and characters with multiple Unicode representations \cite{hasan-etal-2020-low} to avoid [UNK] during tokenization. 
% We generate labels using the normalized annotated texts.

\subsubsection{Class Labelling}
To prepare the dataset for a four-class classification problem (no error(O), begin error(B), inside error(I), missing after(M)), we extract the character indices from the normalized sentence with \$ enclosed annotations corresponding to each class, for example:

\begin{center}
    % {\fontsize{6pt}{14pt}\selectfont BBBB BBB III MMMMM} \\
    % {\beng ডিম ভাজি সাথে \$পুরা\$ \$মাছ টাই\$ খেলাম\$\$}

    % $\overbrace{\text{\beng ডিম ভাজির সাথে}}^{\text{O}} $
    % $\overbrace{\text{\beng \$পুরা\$}}^{\text{B}} $
    % $\overbrace{\text{\beng \$মাছ}}^{\text{B}} $
    % $\overbrace{\text{\beng টাই\$}}^{\text{I}} $
    % $\overbrace{\text{\beng খেলাম\$\$}}^{\text{M}} $
\end{center}

We incorporate a label for the [CLS] token when the proportion of toxic offsets in the text exceeds 30\%. This enables the system to be trained on a proxy text classification objective \cite{chhablani2021nlrg}.

\subsection{Token Classification Models}
% TODO: add model diagrams
\subsubsection{Transformer-based Token Classification Models}
\begin{figure}[h]
  \centering
  % \centerline{\includegraphics[width=0.45\textwidth]{transformers.jpg}}
  \centerline{\includegraphics[width=0.45\textwidth]{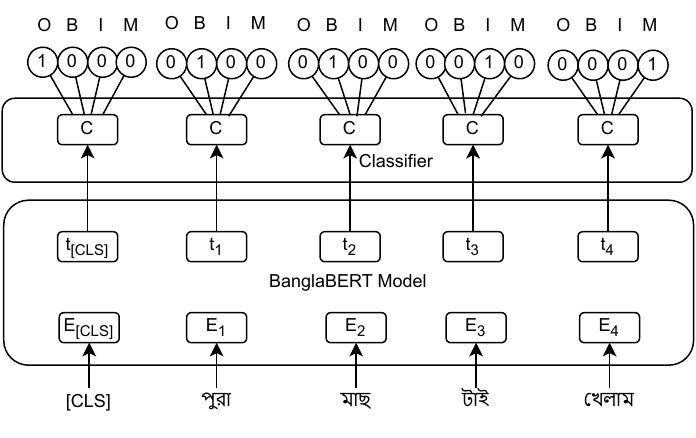}}
  \caption{Transformer-based Token Classification Models}
  \label{fig:model}
\end{figure}

% ডিম ভাজির সাথে পুরা মাছ টাই খেলাম
% ['ডিম', 'ভাজ', '##ির', 'সাথে', 'পুরা', 'মাছ', 'টাই', 'খেলাম']
The Token Classification Model is composed of an ELECTRA-based model, specifically BanglaBERT-base and BanglaBERT-large \cite{bhattacharjee-etal-2022-banglabert}, and a classification layer that is applied to each final token embedding to predict the error class of each token as shown in Figure~\ref{fig:model}.

\subsubsection{LSTM-CRF Token Classification Models}
\begin{figure}[h]
  \centering
  % \centerline{\includegraphics[width=0.45\textwidth]{transformers.jpg}}
  \centerline{\includegraphics[width=0.45\textwidth]{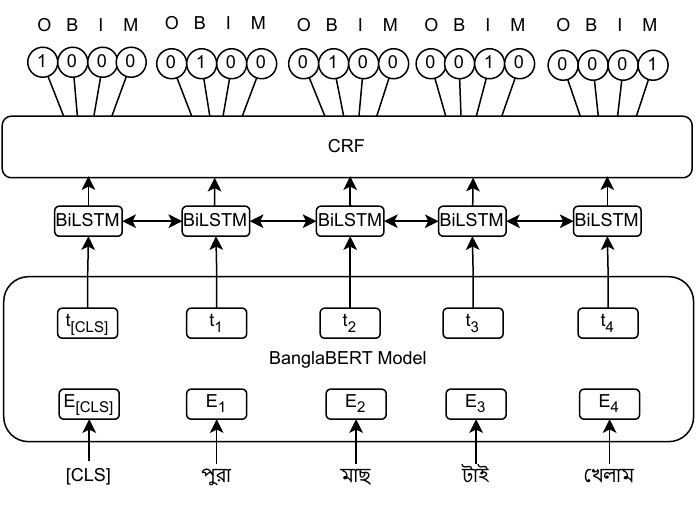}}
  \caption{LSTM-CRF Token Classification Models}
  \label{fig:model2}
\end{figure}

A popular approach for Named-Entity Recognition tasks is to combine Conditional Random Fields (CRF) with transformer-based models, as demonstrated in recent studies \cite{souza2019portuguese,jurkiewicz2020applicaai,chhablani2021nlrg}. In our approach, we utilize BanglaBERT-based models that incorporate a single BiLSTM layer and a CRF layer  as shown in Figure~\ref{fig:model2}.. During training, the CRF loss is applied, while Viterbi Decoding is performed during prediction. 
% While CRF is generally used for word-level classification, we do not mask inner and end tokens. As a result, all tokens belonging to a word are considered for classification.

\subsection{Post-processing}

\subsubsection{Error Span Generation}
% Probability Threshold:
% TODO: add threshold curve

Biasing the method to prefer the original tokens, unless the model is very confident about an error, produces superior results \cite{alikaniotis2019unreasonable}. We choose the confidence threshold that performs best in the dev set, often ranging from 0.7 to 0.8.

Once our model generates the token-level labels, we further process the output to transform it into an array of character offsets that correspond to tokens with error classes. To achieve this, we first map each token to its offset span during tokenization and then extract the character offsets associated with all the error tokens.

\subsubsection{Reverse Normalization}
The models generate error spans with respect to the normalized text, whereas the task expects the final output to be with respect to the text without any modification (such as normalization). To solve this problem, we calculate the minimum Levenshtein edit-distance-based alignment mapping between the normalized and the original texts using \texttt{edit\_distance\_align} function from the NLTK library and apply some rule-based corrections to get the best alignment. Thus, we get the final error spans in the expected form.

\subsection{Ensemble Learning}
Utilizing the predictions from the top few checkpoints and averaging the results helps obtain superior classification scores \cite{chen2017checkpoint,chhablani2021nlrg}. Building on this approach, we also aggregate the predicted spans from different checkpoints within a model, as well as across different models \cite{nguyen2021s}, using union or intersection methods.

\subsection{Deterministic Error Detection}
\subsubsection{Punctuation Error}
If there is any space before periods, commas, question marks, or exclamation marks, we manually mark that span with spaces as an error. Similarly, if the sentence does not end with punctuation, we report a punctuation missing error.

\subsubsection{Spelling Error}
We collect a database of about one million spelling errors from DPCSpell \cite{bijoy2022dpcspell}. We filter out spelling errors that are absent from the online Bangla dictionary. % TODO cite sources of dictionary
Next, to make the models refrain from identifying named entities as spelling errors, we further filter out the spelling errors that are absent from the word collections of the titles of the Bangla Wikipedia pages. Now, we get a filtered collection of one million spelling errors. Finally, we mark a word as an error if the word exits in the filtered spelling error collection and the word is not a named entity. We use the NER model from the open-source BNLP library as a named entity classifier. %TODO cite bnlp

\section{Experiments}

\subsection{Dataset}
We collect the publicly available dataset of Bangla Grammatical Error Detection Challenge~\cite{bengali-ged} that comprises around 25000 texts extracted from various online sources. The training dataset contains a total of around 20000 texts and the test dataset contains 5000 texts. We split the training dataset into train and dev sets for evaluation purposes using an 80:20 split stratified by the number of error spans.

\subsection{Evaluation Metrics}
For this task, we employ the Levenshtein Distance of the predicted string and ground truth to evaluate our models (lower is better). The Levenshtein distance between two strings $a$,$b$ (of length $|a|$ and $|b|$ respectively) is given by ${\displaystyle \operatorname {lev} (a,b)}$ where 

\begin{figure}[h]
  \centering
  \centerline{\includegraphics[width=0.49\textwidth]{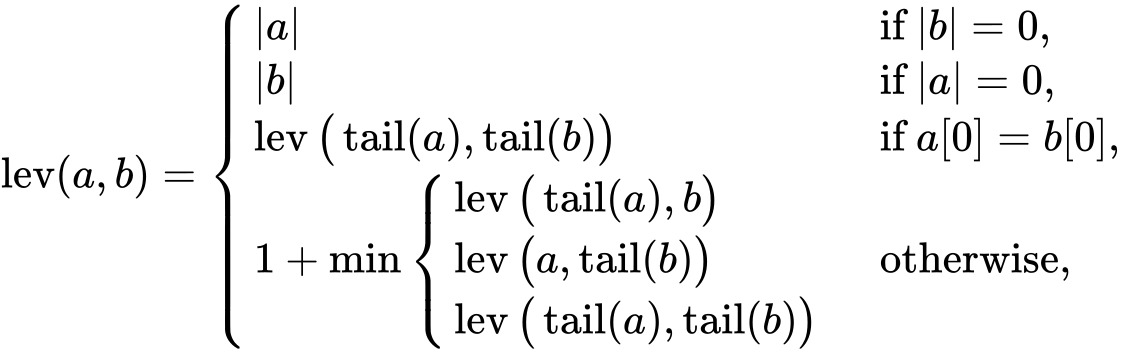}}
  % \caption{Caption goes here.}
  \label{fig:lev}
\end{figure}

\subsection{Implementation Details}
For all pre-trained transformer-based models and tokenizers, we use HuggingFace’s transformers~\cite{wolf2020transformers} in the PyTorch framework.
% TODO add CRF
We finetune BanglaBERT-base and BanglaBERT-large~\cite{bhattacharjee-etal-2022-banglabert} for 30 epochs with batch size of 8. We use an AdamW optimizer~\cite{loshchilov2017decoupled} with learning rate of 2e-5 and weight decay of 0.01. We use a linear learning rate decay with warmup ratio of 0.1. Evaluating the models multiple times during an epoch increases the chance of getting better validation performance~\cite{dodge2020fine}. Hence, we evaluate the model after 500 gradient steps. During tokenization, we use the
maximum length of 384.

% Optimizing the model with cross-entropy loss causes overfitting. 
Label smoothing improves the accuracy of Inception networks on ImageNet~\cite{szegedy2016rethinking} and NLP tasks~\cite{zhu2022boundary,nguyen2021s} by serving as a form of regularization. This involves assigning a small probability to non-ground-truth labels, which can prevent the models from being too confident about their predictions and improve generalization. Label smoothing has proven to be a useful alternative to the standard cross entropy loss and has been widely adopted to address over-confidence~\cite{zoph2018learning, chorowski2016towards, vaswani2017attention}, improve the model calibration \cite{muller2019does}, and de-noise incorrect labels~\cite{lukasik2020does}. 
We use a label smoothing factor of 0.1 for BanglaBERT-base and 0.2 for BanglaBERT-large.

\section{Results}

\subsection{Pre-trained Models}
We initially experiment on different transformers models that are multilingual and for Bangla specifically in order to select the right backbones on a binary classification task of detecting whether a token is an error or not, shown in Table~\ref{tab:backbone-lev}. Here, all the models are trained using standard cross-entropy loss with punctuation post-processing and no normalization. We find that BanglaBERT models are the best candidates for this task.

\begin{table}[htbp]
\centering
\caption{Comparison of Transformers Models on Test Set}
\label{tab:backbone-lev}
\begin{tabular}{lc}
\hline
\textbf{Model} & \textbf{Levenstein Distance} \\ \hline
XLM-RoBERTa-base & 1.394 \\ \hline
DeBERTa-V3-large & 1.3552 \\ \hline
BanglaBERT-base & 1.2120 \\ \hline
BanglaBERT-large & \textbf{1.1844} \\ \hline
% BanglaBERT-base+LSTM+CRF & 1.2036 \\ \hline
% BanglaBERT-large+LSTM+CRF & 1.1488 \\ \hline
\end{tabular}
\end{table}

\subsection{LSTM-CRF}
We do not observe any performance improvement by using LSTM-CRF on top of transformer-based pre-trained models in the dev set. Hence, we  employ BanglaBERT-base and BanglaBERT-large as the final solution.

\subsection{Deterministic Error Detection}
We check for extra spaces before punctuation (space fix) and missing punctuation at the end (end fix). The models fail to capture these errors an we notice performance improvement by applying these modifications on the model output that are presented in Table~\ref{tab:punc-fix}. We also notice a slight improvement in manual spelling error detection using our one million spelling error database. 
\begin{table}[htbp]
\centering
\caption{Effectiveness of deterministic punctuation error detection on the  test set}
\label{tab:punc-fix}
\begin{tabular}{lc}
\hline
\textbf{Model} & \textbf{LD} \\ \hline
BanglaBERT-base & 1.2948 \\ \hline
BanglaBERT-base+space fix & 1.246 \\ \hline
BanglaBERT-base+space fix+end fix & \textbf{1.212} \\ \hline
\end{tabular}
\end{table}

\subsection{Ensemble Strategy}
We explore the effectiveness of span union or intersection methods to ensemble different models as presented in Table~\ref{tab:ens-un-int} and Table~\ref{tab:ens-un-int-multi}. The results suggest that intersection approaches outperform corresponding union and single checkpoint approaches, whereas union approaches perform worse than single checkpoints. These findings imply that the individual checkpoints may be predicting additional offsets that are identified as errors. Hence, We take the intersection of the three best checkpoints of BanglaBERT-base and BanglaBERT-large and combine the two predictions by intersection again to achieve the best Levenstein score in the test set.

\begin{table}[htbp]
\centering
\caption{Effectiveness of ensemble I on test set}
\label{tab:ens-un-int}
\begin{tabular}{lc}
\hline
\textbf{Type} & \textbf{LD} \\ \hline
BanglaBERT-base+large Union & 1.2524 \\ \hline
BanglaBERT-base+large Intersection & \textbf{1.144} \\ \hline
BanglaBERT-large only & 1.2212 \\ \hline
\end{tabular}
\end{table}

\begin{table}[h]
\centering
\caption{Effectiveness of Ensemble II on Test Set}
\label{tab:ens-un-int-multi}
\begin{tabular}{lc}
\hline
\textbf{Model} & \textbf{LD} \\ \hline
Single-checkpoint & 1.0648 \\ \hline
Three-checkpoints & \textbf{1.054} \\ \hline
\end{tabular}
\end{table}

\subsection{Label Smoothing}
Label Smoothing stops the model from overfitting and modeling noise. We find label smoothing to be beneficial to get better performance as shown in Table~\ref{tab:smth}.

\begin{table}[h]
\centering
\caption{Test set results for label smoothing}
\label{tab:smth}
\begin{tabular}{lc}
\hline
\textbf{Type} & \textbf{LD} \\ \hline
BanglaBERT-large+standard CE  & 1.164 \\ \hline
BanglaBERT-large+smoothing 0.2  & 1.1588 \\ \hline
\end{tabular}
\end{table}

\subsection{Thresholding}
\begin{figure}[h]
  \centering
  \centerline{\includegraphics[width=0.48\textwidth]{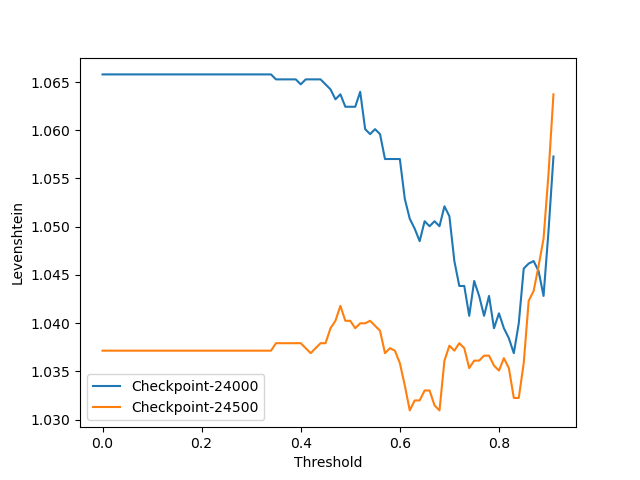}}
  \caption{Effect of confidence thresholding on the dev Set}
  \label{fig:thresh}
\end{figure}

Figure~\ref{fig:thresh} demonstrates the effect of confidence thresholding for two checkpoints of the BanglaBERT model in the dev set. We observe that the model performs best near to a threshold of 0.8, hence we choose this threshold for BanglaBERT models during inference which boosts the test set performance as shown in Table~\ref{tab:thresh}.

\begin{table}[h]
\centering
\caption{Results of thresholding on the test set}
\label{tab:thresh}
\begin{tabular}{lc}
\hline
\textbf{Type} & \textbf{LD} \\ \hline
BanglaBERT-large+threshold 0.0  & 1.1892 \\ \hline
BanglaBERT-large+threshold 0.8  & \textbf{1.1588} \\ \hline
\end{tabular}
\end{table}

\subsection{Unicode Normalization}
Table~\ref{tab:norm} shows the effectiveness of union normalization. We observe a performance boost when we use the normalization and the reverse normalization properly. For all the models, we use this scheme.

\begin{table}[h]
\centering
\caption{Results of Unicode normalization on the test set}
\label{tab:norm}
\begin{tabular}{lc}
\hline
\textbf{Type} & \textbf{LD} \\ \hline
BanglaBERT-large without normalization  & 1.130 \\ \hline
BanglaBERT-large with normalization  & \textbf{1.084} \\ \hline
\end{tabular}
\end{table}

\subsection{The Final System}
The final system is an intersection ensemble of the three best checkpoints of BanglaBERT-base and BanglaBERT-large. We apply all the aforementioned techniques with the hyper-parameters that perform best in the dev set and the test set.

\section{Discussion}

We generate synthetic datasets from the Prothom-Alo scrapes available online to simulate real grammar errors~\cite{rahman2022bspell}. Though training with this additional data slightly improves the dev set metric, it hurts the test set performance. The cause for this contrast could be attributed to the inconsistent distribution of data in the dev and test sets.

The dataset size is small with respect to the model complexity of the transformer models. For this reason, overfitting occurs easily if we do not apply label smoothing, thresholding, and learning rate scheduling to regularize and help the model reach a generalized solution.

Previous works~\cite{wang2021hitmi,ghosh2021cisco} report performance boost by employing LSTM-CRF with transformer-based models in span detection tasks, but in contrast we find vanilla transformer-based models to perform better. This can happen as we focus on tuning the hyper-parameters for different proposed modifications with respect to vanilla transformer-based models only. Extensive hyper-parameter search and optimization strategies can potentially make LSTM-CRF augmented models demonstrate their full potential.

\begin{figure}[h]
  \centering
  % \centerline{\includegraphics[width=0.45\textwidth]{transformers.jpg}}
  \centerline{\includegraphics[width=0.45\textwidth]{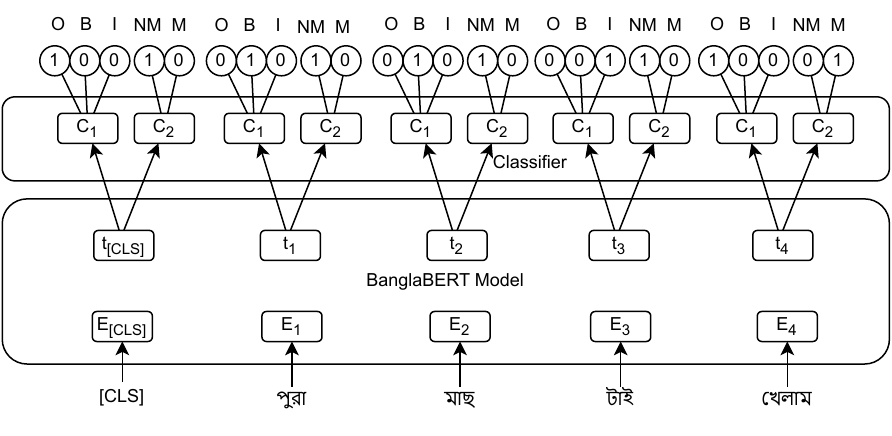}}
  \caption{Separate head for missing errors}
  \label{fig:model3}
\end{figure}

To model the missing errors (empty spans after a character), an alternative approach can be using character-level embedding instead of token-level embedding and formalizing this type of error as a separate classification problem apart from non-empty spans detection as shown in Figure~\ref{fig:model3}. We have used a separate classification head for this error but do not observe much improvement.

\section{Conclusion}
% In this work, we present our approach to Bhashabhrom: Bangla Grammatical Error Detection Challenge. We achieve the best scores in the public and private test set leaderboards using a transformer-based ensembles with a bag of tricks. 

In this work, we present our approach to solving the Bangla grammatical error detection problem that outperforms multiple baselines. We formalize the problem as a sequence labeling problem and apply different transformer-based models and feature-based models. BanglaBERT, an ELECTRA model pre-trained on clean Bangla corpus, is at the heart of our system. We also find that clever choice of ensemble, loss function, and rule-based post-processing significantly improves the machine learning-based systems. Our model is effective in detecting various kinds of grammar errors in Bangla and will motivate deep learning-based approaches to solve this complex problem.

In future,  we intend to enhance our model by employing self-training with in-domain unlabeled data, combining self-training with feature-based learning to learn a more robust model, and using Fast Gradient Method (FGM) as an adversarial training strategy. Moreover, we will evaluate our system for the Bangla grammar error correction problem.

\bibliographystyle{acl_natbib}
\bibliography{custom}

\end{document}